\documentclass[conference]{IEEEtran}
\IEEEoverridecommandlockouts
\usepackage{cite}
\usepackage{amsmath,amssymb,amsfonts}
\usepackage{algorithmic}
\usepackage{graphicx}
\usepackage{textcomp}
\usepackage{xcolor}
\usepackage{enumitem}
\usepackage{booktabs,multirow}
\IEEEoverridecommandlockouts
\usepackage{ragged2e}
\usepackage{braket}
\def\BibTeX{{\rm B\kern-.05em{\sc i\kern-.025em b}\kern-.08em
    T\kern-.1667em\lower.7ex\hbox{E}\kern-.125emX}}

\usepackage{tikz}
\usetikzlibrary{svg.path}
\usepackage[hidelinks]{hyperref}
\usepackage{scalerel}

\definecolor{orcidlogocol}{HTML}{A6CE39}
\tikzset{
  orcidlogo/.pic={
    \fill[orcidlogocol] svg{M256,128c0,70.7-57.3,128-128,128C57.3,256,0,198.7,0,128C0,57.3,57.3,0,128,0C198.7,0,256,57.3,256,128z};
    \fill[white] svg{M86.3,186.2H70.9V79.1h15.4v48.4V186.2z}
                 svg{M108.9,79.1h41.6c39.6,0,57,28.3,57,53.6c0,27.5-21.5,53.6-56.8,53.6h-41.8V79.1z M124.3,172.4h24.5c34.9,0,42.9-26.5,42.9-39.7c0-21.5-13.7-39.7-43.7-39.7h-23.7V172.4z}
                 svg{M88.7,56.8c0,5.5-4.5,10.1-10.1,10.1c-5.6,0-10.1-4.6-10.1-10.1c0-5.6,4.5-10.1,10.1-10.1C84.2,46.7,88.7,51.3,88.7,56.8z};
  }
}

\newcommand\orcidicon[1]{\href{https://orcid.org/#1}{\mbox{\scalerel*{
\begin{tikzpicture}[yscale=-1,transform shape]
\pic{orcidlogo};
\end{tikzpicture}
}{|}}}}

\begin{document}

\title{Battery health prognosis using Physics-informed neural network with Quantum Feature mapping}

\author{Muhammad~Imran~Hossain$^{\textsuperscript{\orcidicon{0009-0001-2155-1421}}}$\,,~\IEEEmembership{Graduate Student Member,~IEEE}
, Md~Fazley~Rafy$^{\textsuperscript{\orcidicon{0000-0003-3057-9546}}}$\,,~\IEEEmembership{Member, IEEE},\\Sarika~Khushalani~Solanki,~\IEEEmembership{Senior~Member,~IEEE}, ~Anurag~K.~Srivastava$^{\textsuperscript{\orcidicon{0000-0003-3518-8018}}}$\,,~\IEEEmembership{Fellow,~IEEE}
\thanks{Authors are with the Lane Department of Computer Science and Electrical Engineering, West Virginia University, Morgantown, WV, USA, 26505. Authors would like to acknowledge support from ARC.}}
\maketitle

\IEEEpubid{\makebox[\columnwidth]{979-8-3315-5720-1/26/\$31.00 ©2026 IEEE\hfill}\hspace{\columnsep}\makebox[\columnwidth]{}} 
\IEEEpubidadjcol

\begin{abstract}
Accurate battery health prognosis using State of Health (SOH) estimation is essential for the reliability of multi-scale battery energy storage, yet existing methods are limited in generalizability across diverse battery chemistries and operating conditions. The inability of standard neural networks to capture the complex, high-dimensional physics of battery degradation is a major contributor to these limitations. To address this, a physics-informed neural network with the Quantum Feature Mapping(QFM) technique (QPINN) is proposed. QPINN projects raw battery sensor data into a high-dimensional Hilbert space, creating a highly expressive feature set that effectively captures subtle, non-linear degradation patterns using Nystr\"om method. These quantum-enhanced features are then processed by a physics-informed network that enforces physical constraints. The proposed method achieves an average SOH estimation accuracy of 99.46\% across different datasets, substantially outperforming state-of-the-art baselines, with reductions in MAPE and RMSE of up to 65\% and 62\%, respectively. This method was validated on a large-scale, multi-chemistry dataset of 310,705 samples from 387 cells, and further showed notable adaptability in cross-validation settings, successfully transferring from one chemistry to another without relying on target-domain SOH labels.
\end{abstract}


\section{Introduction}\label{sec:intro}

\IEEEPARstart{D}{espite} its importance, accurate State of Health (SoH) estimation of grid-scale battery energy storage faces significant challenges. It is persistent due primarily to the fact that existing data-driven methods rely on specific electrochemical models that limit generalization across diverse battery chemistries \cite{MU2024110221}\cite{pr13113559} \cite{math11204263}. This challenge is compounded by the limited representational capacity of standard neural network components within Physics-Informed Neural Networks (PINNs), which often struggle to capture the complex fully, high-dimensional relationships inherent in battery degradation physics \cite{wang2024physics}. 

Empirical and model-based methods (e.g., Kalman filters with ECMs) have practical limits for SOH estimation. The Dual Extended Kalman Filter \cite{9935824}, though rigorous, depends on fixed circuit models that need offline re-identification as conditions change. Its sensitivity to initialization, tuning, and model mismatch, plus limited testing across calendar aging, diverse operating profiles, and complex degradation modes, reduces its robustness for real-world deployments \cite{MU2024110221} ,\cite{8916734}. On the contrary, data-driven methods (LSTM, GRU, BRNN) offer flexibility and a strong fit to benchmark datasets but pose significant challenges\cite{pr13113559}. As black-box models, they depend on phenomenological features and lack physical interpretability \cite{9162518}, \cite{10506507}. Advanced architectures like GWO-BRNN and tuned GRU excel at curve fitting yet are prone to dataset-specific overfitting and poor generalization across chemistries, temperatures, and usage patterns \cite{math11204263}. Recently, PINNs have embedded governing equations into learning and improved monotonicity and smoothness. However, current SOH implementations still rely on hand-crafted statistical features (means, slopes, incremental capacities) that may miss high-order correlations in voltage and temperature traces \cite{pr13113559}, \cite{10251604}, \cite{TIAN2025137215}. As a result, PINNs can perform well in-sample yet struggle to generalize across chemistries and irregular operating conditions. Electrode-level feature methods improve interpretability \cite{wang2024physics} but add computational cost and remain sensitive to electrochemical model mismatch, limiting robustness in practical deployments. Quantum-inspired methods show conceptual promise for battery state-of-health estimation but remain nascent. Current quantum frameworks mainly use it as metaphors for density-based clustering, rely on generic low-level features, and lack electrochemistry-specific adaptation or integration with physical degradation models \cite{7328705}.
\begin{figure*}[htb!]
    \centering
\includegraphics[width=1\linewidth]{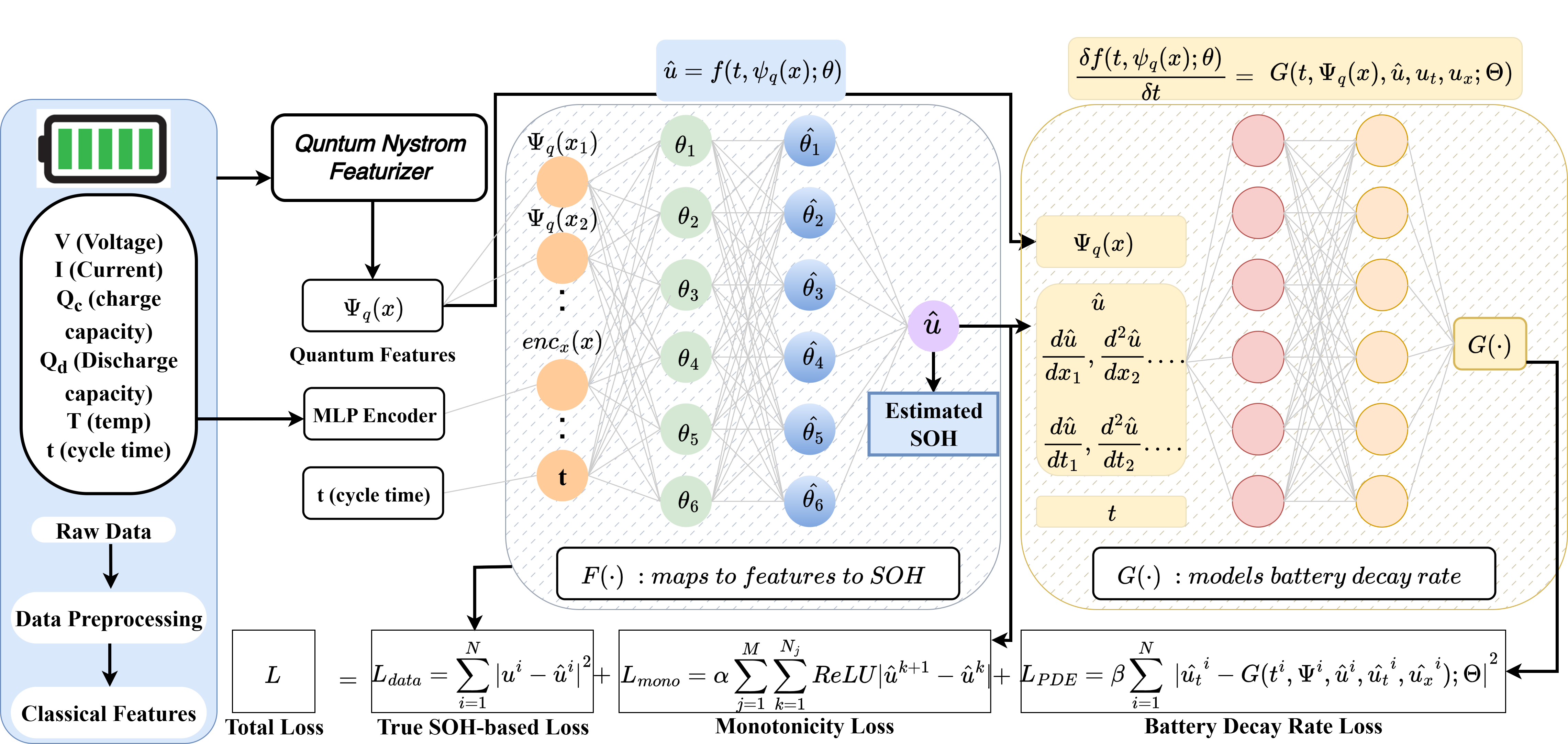}
    \caption{Model Architecture for Physics-informed Neural Network with quantum feature mapping (Nyström-approximated)}
    \label{fig:model_arch}
\end{figure*}

\noindent Recent progress in battery SOH estimation has highlighted two persistent gaps across data-driven and physics-guided approaches. Classical regression models and standard deep-learning approaches \cite{10251604}, \cite{TIAN2025137215} can capture overall aging trends, but they often fail to generalize across different battery chemistries. Their reliance on fixed, low-dimensional features limits their ability to represent the complex and nonlinear behaviors present in real charge–discharge cycles~\cite{havlivcek2019supervised, suzuki2020analysis}. Physics-informed methods improve consistency by embedding degradation dynamics, yet their performance is highly sensitive to the quality and structure of the input features, which often remain too coarse or noisy to support stable training across diverse operating conditions. Motivated by these limitations, prior studies on Hilbert-space feature mappings and quantum-inspired kernels~\cite{schuld2019quantum, jerbi2023quantum} demonstrate that richer, geometry-aware representations can substantially enhance separability and robustness. This conceptual evolution naturally leads to the QFM–PINN framework proposed in this work.

\subsection{Key Contributions}\label{sec:contribution}
\noindent To address the limitations of the traditional battery SoH estimation, we proposed the Quantum Feature Mapping (QFM) in the PINN architecture as shown in Fig. \ref{fig:model_arch}. The key contributions are summarized below.
\begin{itemize}[leftmargin=*]
    \item Introduced a dedicated quantum-inspired kernel-based feature extraction layer to project classical battery sensor data (voltage, current, temperature) into a rich, high-dimensional Hilbert space. This quantum embedding was shown to capture complex, non-linear relationships and degradation patterns that are challenging for classical neural networks to learn, providing a more expressive feature set that directly enhances the physics-informed learning process and leads to superior SOH estimation accuracy.
    \item Developed the quantum-feature-based physics-informed neural network (QPINN) to estimate battery SOH solely from voltage–current–temperature charge profiles and their statistical descriptors. The method was designed to be battery model-independent because it does not rely on any predefined electrochemical or equivalent-circuit model. Although the features were shown to implicitly reflect characteristics of the underlying chemistry, the architecture was made adaptable to other battery types pending appropriate validation. Unlike classical statistical features, QFM produces expressive representations that highlight non-linear dependencies across cycles and chemistries, enabling the PINN to learn degradation patterns that are otherwise difficult to capture.
    \item Performed extensive validation on 310,705 samples in 387 batteries with four different chemistries, and QPINN was shown to demonstrate superior accuracy over both baseline physics-informed networks and KAN-based variants, both in validation (Table \ref{Tab.1}) and cross-validation (Table \ref{Tab.2}) as described in Section \ref{sec:results}.
\end{itemize}
\begin{figure}
    \centering
    \includegraphics[width=1\linewidth]{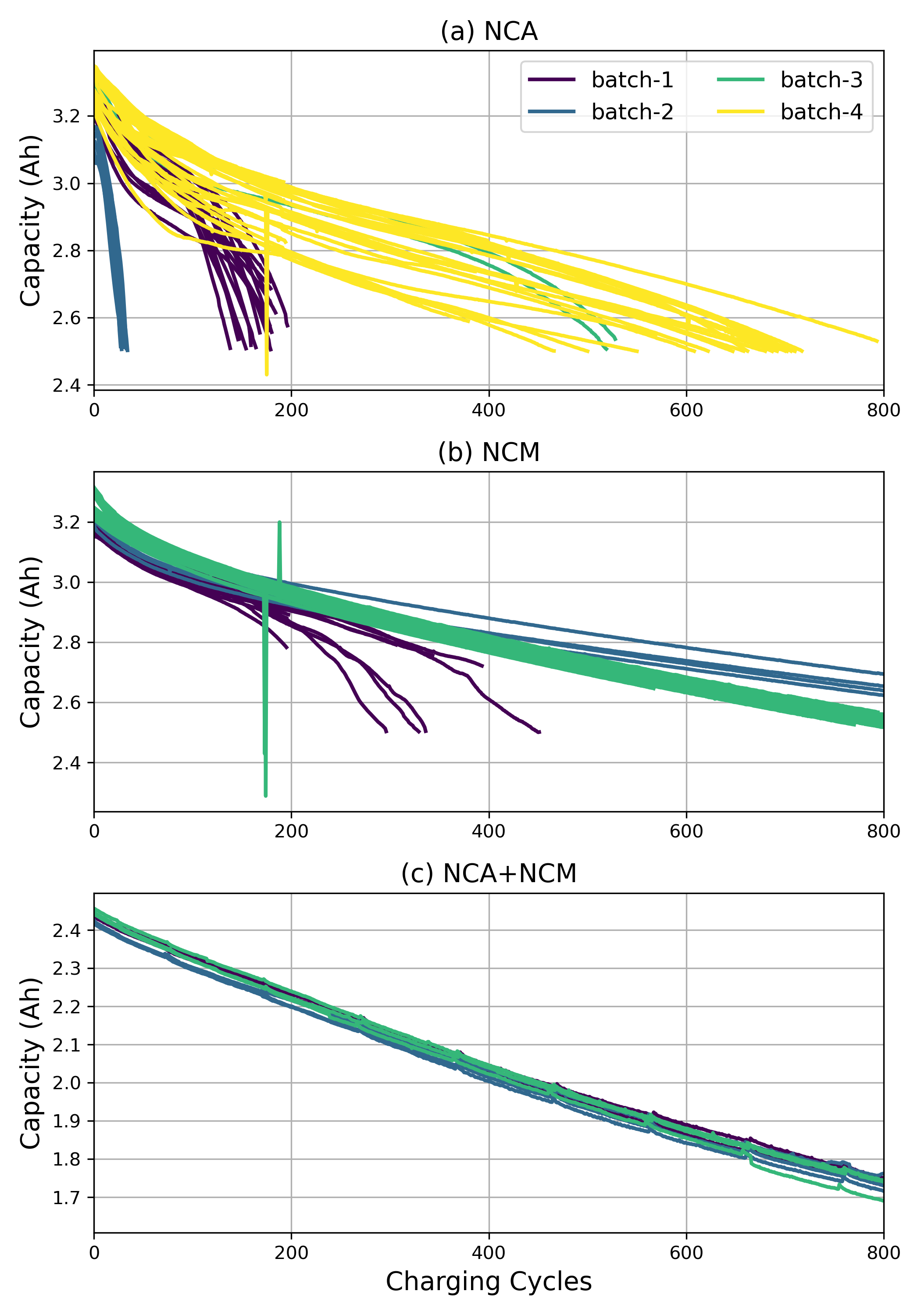}
    \caption{Degradation patterns in the TJU \cite{Zhu2022_Zenodo} battery dataset, comprising four batches (130 batteries in total) and three different chemistries (NCA, NCM, and NCA-NCM). Charge/discharge protocols vary across batches.}
    \label{fig:degrad_pat}
\end{figure}

\section{Proposed Methodology of QPINN}\label{sec:methodology}

The proposed QPINN framework addresses the limitations of existing SOH estimation approaches through a novel integration of quantum-inspired feature extraction and physics-constrained learning. As illustrated in Fig. \ref{fig:model_arch}, our methodology comprises three interconnected components that work collaboratively to enable robust, generalizable battery health prognosis. The overall architecture begins with raw battery operational data (voltage, current, temperature time series), which transforms into a quantum feature extraction layer to generate rich, high-dimensional representations in Hilbert space. These quantum-enhanced features are then fed into a feed-forward neural network model and into a physics-informed neural network that simultaneously learns from data while respecting fundamental electrochemical constraints. Finally, the framework outputs accurate SOH estimates with inherent physical constraints. The proposed QPINN augments a baseline PINN~\cite{wang2024physics}, as explained in Section \ref{sec:pinn_baseline}, with a fixed, low-rank quantum kernel embedding (Section \ref{sec:qfeat}), yielding a hybrid input geometry that improves optimization stability, cross-dataset generalization, and interpretability.

\subsection{Baseline PINN}
\label{sec:pinn_baseline}

\noindent Let $x \in \mathbb{R}^d$ denote cycle descriptors (e.g., statistics of voltage/current/temperature) and $t \in \mathbb{R}$ denote cycle/time. The solution network models a continuous SoH field whose temporal evolution is constrained by an auxiliary dynamics network $G(\cdot)$ through the physics residual. The residual $H(t,x)$ in Eq.~\eqref{eq:resid} enforces that the learned solution (Eq. \eqref{eq:sol}) obeys degradation dynamics, where $u_t \!=\! \partial u/\partial t$ and $u_x \!=\! \partial u/\partial x$.
\begin{equation}
u \;=\; F(t,x;\,\Phi),
\label{eq:sol}
\end{equation}
\begin{equation}
H(t,x) \;=\; u_t \;-\; G\!\big(t,x, u, u_t, u_x;\,\Theta\big)
\label{eq:resid}
\end{equation}
\noindent Training minimizes a composite objective consisting of (i) a data-consistency term, (ii) the physics residual~\eqref{eq:resid}, and (iii) a monotonicity regularizer to avoid non-physical SoH increases as shown in Eq. ~\eqref{eq:loss}, which directly suppresses increases of $u$ across successive cycles (the penalty~\eqref{eq:mono} contributes to~\eqref{eq:loss}). For the monotonicity term we use a forward-difference penalty along $t$ in Eq. \eqref{eq:mono}.
\begin{equation}
L \;=\; L_{\text{data}} \;+\; \alpha\, L_{\text{PDE}} \;+\; \beta\, L_{\text{mono}}.
\label{eq:loss}
\end{equation}
\begin{equation}
L_{\text{mono}} \;=\; \frac{1}{N}\sum_{k=1}^{N-1} 
\Big(\max\{0,\, u(t_{k+1},x_k)-u(t_k,x_k)\}\Big)^{2}
\label{eq:mono}
\end{equation}

\subsection{Quantum Feature Mapping}
\label{sec:qfeat}

To precondition the input geometry, we embed $x$ in a $n$-qubit Hilbert space using a data-dependent unitary $\mathcal{U}_\phi(x)$~\cite{schuld2019quantum, havlivcek2019supervised} as shown in Eq. \eqref{eq:qstate}.
\begin{equation}
\ket{\psi(x)} \;=\; \mathcal{U}_\phi(x)\,\ket{0}^{\otimes n}.
\label{eq:qstate}
\end{equation}
The induced quantum kernel compares the inputs via the state overlap in Eq. \eqref{eq:qkernel}.
\begin{equation}
K(x_i,x_j) \;=\; \big|\langle \psi(x_i)\,|\,\psi(x_j)\rangle\big|^{2}.
\label{eq:qkernel}
\end{equation}
Following~\cite{havlivcek2019supervised, suzuki2020analysis}, we control expressivity through feature-map depth and entangling structures, after which we implement data re-uploading to enrich interactions~\cite{P_rez_Salinas_2020}. For scalability, the kernel feature map is approximated via Nystr\"om landmarks $S=\{s_m\}_{m=1}^{M}$. The resulting low-rank embedding produces an orthonormalized $M$-dimensional representation (the map in Eq. ~\eqref{eq:nystrom} is used as a fixed encoder in our model). As argued in recent analyses~\cite{jerbi2023quantum}, quantum kernels can compactly capture high-order feature interactions, offering a richer geometry than standard classical kernels in certain regimes.
\begin{equation}
\psi_q(x) \;=\; K(x,S)\,K(S,S)^{-1/2}
\label{eq:nystrom}
\end{equation}

\subsection{Proposed QPINN}
\label{sec:qpinn}

QPINN preserves the integration of the PINN formulation of Eq.~\eqref{eq:sol}--\eqref{eq:loss} but replaces the raw input with a \emph{hybrid} representation that concatenates a fixed quantum embedding with a small trainable encoder (Eq.~\eqref{eq:hybrid}), where the solution and dynamics networks operate on $z$:
\begin{equation}
z(t,x) \;=\; \big[\,\psi_q(x)\,,\; \mathrm{enc}_x(x;\,\Xi)\,,\; t\,\big].
\label{eq:hybrid}
\end{equation}
The solution and dynamics networks then act on $z$ as
\begin{align}
u \;&=\; F\!\big(z(t,x);\;\Phi\big),
\label{eq:qsolve}
\\[2pt]
H(t,x) \;&=\; u_t \;-\; G\!\big(z(t,x),\,u,\,u_t,\,u_x;\;\Theta\big),
\label{eq:qdyn}
\end{align}
and QPINN is trained by minimizing the same objective in Eq.~\eqref{eq:loss} with $L_{\text{mono}}$ in Eq.~\eqref{eq:mono}. 
Crucially, $\psi_q(\cdot)$ in Eq.~\eqref{eq:hybrid} is \emph{fixed} (stop-gradient), while $\mathrm{enc}_x(\cdot)$ remains trainable to preserve a differentiable path from $x$ to $u$ so that $u_x$ in Eq.~\eqref{eq:qdyn} is well-defined.

 QPINN simultaneously performs feature-level and model-level information fusion~\cite{wen2023physics}. At the \emph{feature level}, heterogeneous descriptors of the battery operating condition encoded in $x$ are embedded into a joint representation $z(t,x)$ that merges the fixed quantum feature map $\psi_q(x)$ with the trainable encoder output $\mathrm{enc}_x(x;\Xi)$ and temporal information $t$. At the \emph{model level}, this hybrid representation is coupled with physics-based degradation dynamics through the residual $H(t,x)$ and the composite loss in Eq.~\eqref{eq:loss}, integrating data-driven learning and prior physical knowledge within a single unified architecture. 

 The fixed, well-conditioned basis $\psi_q(x)$ in Eq.~\eqref{eq:nystrom} provides a stable feature map that pre-separates degradation modes and reduces variance during training. The lightweight encoder $\mathrm{enc}_x(\cdot)$ supplies only the flexibility needed to recover $u_x$ and align local statistics with the residual term in Eq.~\eqref{eq:qdyn}, so that the PDE penalty acts on a smoother function class over $x$ and Eq.~\eqref{eq:resid} serves as a stabilizing regularizer rather than a competing objective. Since the Nystr\"om map in Eq.~\eqref{eq:nystrom} approximates the RKHS induced by Eq.~\eqref{eq:qkernel}, training Eqs.~\eqref{eq:qsolve}--\eqref{eq:qdyn} on these fixed features is effectively equivalent to learning a shallow model in that RKHS with physics-informed regularization through Eqs.~\eqref{eq:resid} and \eqref{eq:mono}, decoupling representation and dynamics, and enforcing a geometric inductive bias that improves the bias-variance trade-off.

\begin{figure}
    \centering
    \includegraphics[width=1\linewidth]{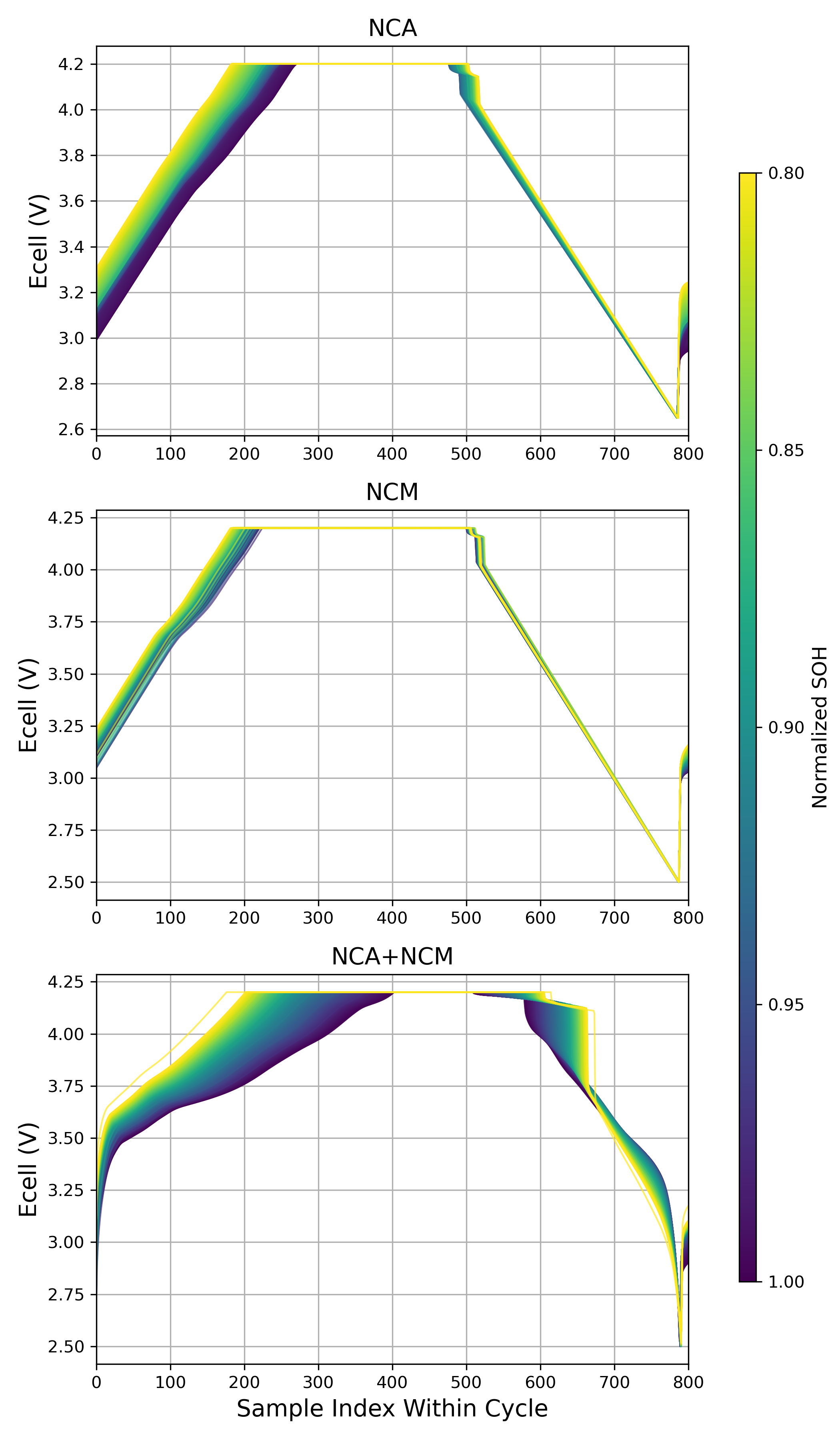}
    \caption{Normalized Voltage Aging Profiles of Three Commercial 18650 Cells (NCA, NCM, and NCM+NCA) from TJU dataset \cite{Zhu2022_Zenodo}}
    \label{fig:aging_pat}
\end{figure}

\section{Datasets and Case Study}
\label{sec:expset}
We evaluate the proposed QPINN framework using four publicly available Li-ion datasets covering LFP, LCO, NCA, and NMC chemistries~\cite{wang2024physics}.  
Each dataset contains a complete cycle-level time series of terminal voltage, current, temperature, and measured capacity, enabling consistent feature extraction and providing ground-truth SOH trajectories. Fig. \ref{fig:degrad_pat} highlights the variation in charge/discharge protocols and the resulting capacity fade trajectories used in our evaluation. The normalized voltage aging profiles for three commercial 18650 cells (NCA, NCM, NCM+NCA) from the TJU dataset are shown in Fig. \ref{fig:aging_pat}, representing the characteristic voltage curve shifts over their battery lifespan. These profiles provide insight into the chemistry-specific electrochemical changes that the model features must capture. All experiments were conducted on a workstation equipped with an AMD Ryzen\textsuperscript{\textregistered}~7 7700X (8 cores, 4.50\,GHz), 32\,GB RAM, and an NVIDIA\textsuperscript{\textregistered} GeForce RTX\texttrademark{}~5070\,Ti GPU with 16\,GB VRAM, running Windows~11 and Python~3.10.
For source-domain training, QPINN was optimized for 300~epochs using the Adam optimizer with an initial learning rate of $1\times10^{-3}$.  
The learning rate was reduced by a factor of 0.1 after 50~epochs without improvement in the validation loss.  
Weight decay of $1\times10^{-5}$ and gradient clipping with a threshold of 1.0 were employed to enhance numerical stability and generalization.

\noindent The quantum feature map used $n{=}8$ qubits and the depth of the circuit $L{=}2$, which provided the best balance between expressivity and numerical stability. An 8-qubit embedding produces a $2^{8}{=}256$-dimensional Hilbert space that matches the Nystr\"om landmark size ($M{=}256$), ensuring a full-rank kernel without the instability or computational overhead of larger quantum embeddings. A depth of $L{=}2$ was sufficient to generate meaningful nonlinear entanglement while avoiding the gradient degradation and poor kernel conditioning observed with deeper circuits. The loss weights $(\alpha,\beta)=(0.7,0.2)$ were selected through coarse grid search, where larger $\alpha$ strengthened physical consistency but slowed optimization, and larger $\beta$ over-penalized monotonicity and reduced accuracy. These settings consistently yielded stable convergence and the lowest validation errors across all datasets.

\noindent Cross-domain fine-tuning was performed for 100~epochs at a learning rate of $5\times10^{-4}$, during which only the encoder and solution network were updated while quantum features remained detached, and the dynamics network was kept frozen.

\begin{table*}[t]
\centering
\caption{The results of proposed QPINN (Ours), physics-informed neural network (PINN), multi-layer perceptron (MLP), convolutional neural network (CNN), and SPIKAN on four datasets.}
\renewcommand{\arraystretch}{1.25}
\begin{tabular}{l c | cc | cc | cc | cc | cc}
\toprule
\textbf{Dataset} & \textbf{Batch} 
& \multicolumn{2}{c|}{\textbf{Ours (QPINN)}} 
& \multicolumn{2}{c|}{\textbf{PINN}}
& \multicolumn{2}{c|}{\textbf{MLP}}
& \multicolumn{2}{c|}{\textbf{CNN}}
& \multicolumn{2}{c}{\textbf{SPIKAN}} \\
\cmidrule(lr){3-4} 
\cmidrule(lr){5-6}
\cmidrule(lr){7-8}
\cmidrule(lr){9-10}
\cmidrule(lr){11-12}
& & MAPE & RMSE & MAPE & RMSE & MAPE & RMSE & MAPE & RMSE & MAPE & RMSE \\
\midrule

\multirow{6}{*}{\textbf{XJTU}}
& 1 & \textbf{0.0026} & \textbf{0.0036} & 0.0070 & 0.0094 & 0.0260 & 0.0277 & 0.0270 & 0.0330 & 0.0422 & 0.0438 \\
& 2 & \textbf{0.0049} & \textbf{0.0062} & 0.0113 & 0.0122 & 0.0275 & 0.0304 & 0.0298 & 0.0352 & 0.0628 & 0.0655 \\
& 3 & \textbf{0.0061} & \textbf{0.0077} & 0.0086 & 0.0100 & 0.0211 & 0.0237 & 0.0177 & 0.0212 & 0.0460 & 0.0383 \\
& 4 & \textbf{0.0039} & \textbf{0.0053} & 0.0071 & 0.0105 & 0.0200 & 0.0234 & 0.0150 & 0.0189 & 0.0309 & 0.0352 \\
& 5 & \textbf{0.0044} & \textbf{0.0060} & 0.0105 & 0.0135 & 0.0183 & 0.0217 & 0.0350 & 0.0453 & 0.0624 & 0.0645 \\
& 6 & \textbf{0.0069} & \textbf{0.0083} & 0.0072& 0.0097 & 0.0204 & 0.0242 & 0.0149 & 0.0194 & 0.0583 & 0.0644 \\
\midrule

\multirow{3}{*}{\textbf{TJU}}
& 1 & \textbf{0.0137} & \textbf{0.0157} & 0.0164 & 0.0158 & 0.0206 & 0.0197 & 0.0198 & 0.0208 & 0.0244 & 0.0241 \\
& 2 & \textbf{0.0093} & \textbf{0.0109} & 0.0119 & 0.0132 & 0.0149 & 0.0157 & 0.0143 & 0.0149 & 0.0324 & 0.0296 \\
& 3 & \textbf{0.0043} & \textbf{0.0044} & 0.0080 & 0.0079 & 0.0150 & 0.0144 & 0.0124 & 0.0125 & 0.0078 & 0.0067 \\
\midrule

\textbf{MIT}
& -- & \textbf{0.0050} & \textbf{0.0067} & 0.0065 & 0.0074 & 0.0079 & 0.0087 & 0.0065 & 0.0075 & 0.0482 & 0.0536 \\
\midrule

\textbf{HUST}
& -- & \textbf{0.0026} & \textbf{0.0035} & 0.0078 & 0.0087 & 0.0080 & 0.0090 & 0.0074 & 0.0087 & 0.0089 & 0.0105 \\
\bottomrule
\end{tabular}

\vspace{2mm}
\footnotesize
MAPE is the mean absolute percentage error, and RMSE is the root mean square error. 
The best results (QPINN) are shown in bold. All values are averages of 10 experiments.
\label{Tab.1}
\end{table*}

\subsection{Baseline Model Comparison}\label{sec:datacom}
To contextualize the performance of QPINN, we compared it against four representative architectures:  
(i) a standard PINN\cite{wang2024physics},  
(ii) a Kolmogorov--Arnold Network–based PINN (PIKAN)\cite{zhang2025physics},  
(iii) a conventional multilayer perceptron (MLP) \cite{10752534}, and  
(iv) a convolutional neural network (CNN) \cite{zhang2022prognostics}.  
In PIKAN, the nonlinear layers were replaced with Kolmogorov–Arnold operators to provide a richer functional approximation of degradation dynamics.  
We additionally evaluated the separable physics-informed KAN model (SPIKAN), which introduces a structured decomposition of the degradation operator under identical training protocols.
\subsection{Cross-Dataset Testing} \label{sec:cross}
For transfer experiments, each model was first trained on a designated source dataset for 200~epochs and subsequently fine-tuned on a target dataset for 100~epochs. During fine-tuning, the physics-based dynamics network $G$ was held fixed to retain the structural prior learned from the source domain, while the solution network $F$ was updated to adapt to the target chemistry. Quantum kernel features remained non-trainable throughout, ensuring that adaptation occurred strictly through the classical pathway and preserving the stability of the hybrid physics-informed formulation. For cross-dataset adaptation, we freeze the dynamics parameters $\Theta$ in Eq. ~\eqref{eq:qdyn} and fine-tune only the solution-side parameters $\Phi$ and the encoder $\Xi$ on the target dataset. This preserves the learned structural prior encoded by $G(\cdot)$ while allowing a compact adapter (\(\mathrm{enc}_x\) and the readout in \(F\)) to realign target statistics within the stable geometry provided by the Eq.~\eqref{eq:nystrom} and Eq. ~\eqref{eq:hybrid}.

\begin{table}[t]
\centering
\scriptsize
\setlength{\tabcolsep}{4pt} 
\caption{Cross-dataset RMSE comparison of QPINN under source-only training and fine-tuning with the dynamics network $G$ frozen.}
\begin{tabular}{l|cccc|cccc}
\toprule
& \multicolumn{4}{c|}{\textbf{Fine-tuning}} 
& \multicolumn{4}{c}{\textbf{Source-only}} \\
\cmidrule(lr){2-5} \cmidrule(lr){6-9}
& XJTU & TJU & MIT & HUST 
& XJTU & TJU & MIT & HUST \\
\midrule
XJTU & -- & \underline{0.0181} & \textit{0.0112} & \textbf{0.0069} 
      & 0.0089 & 0.0737 & 0.0349 & 0.1298 \\
TJU  & \textbf{0.0056} & -- & \textit{0.0109} & \underline{0.0185} 
      & 0.0866 & 0.0110 & 0.1101 & 0.1823 \\
MIT  & \textit{0.0059} & \underline{0.0162} & -- & \textbf{0.0041} 
      & 0.0567 & 0.1177 & 0.0093 & 0.0783 \\
HUST & \textit{0.0052} & \underline{0.0164} & \textbf{0.0051} & -- 
      & 0.0413 & 0.1551 & 0.0560 & 0.0082 \\
\bottomrule
\end{tabular}
\begin{flushleft}
\footnotesize
\justify{The top 3 results are in bold, italic, and underlined, respectively. Each row corresponds to the dataset used to train the QPINN model (source domain), and each column corresponds to the dataset on which the model is evaluated (target domain).}
\end{flushleft}
\label{Tab.2}
\end{table}

\section{Results and Discussion}\label{sec:results}

The quantum featureization contributes richer nonlinear separability, while the physics-informed structure ensures stable extrapolation across cycles, chemistries, and operating regimes. Table~\ref{Tab.1} reports the state-of-health (SoH) estimation accuracy of the proposed QPINN compared with four representative baseline models: PINN, MLP, CNN, and SPIKAN. Across all datasets and evaluation batches, QPINN consistently achieves the lowest MAPE and RMSE, demonstrating clear superiority in both absolute accuracy and robustness. On the XJTU dataset, which contains six operating batches with varying degradation behaviors, QPINN outperforms the best baseline by substantial margins. The improvements in MAPE range from \textbf{4\% to 63\%}, while RMSE reductions fall between \textbf{14\% and 62\%}. These results highlight QPINN’s ability to model complex cycle-to-cycle degradation dynamics more effectively than purely data-driven or classical physics-informed architectures. Similarly strong performance is observed on the TJU dataset. QPINN improves over the closest baseline by \textbf{16–45\%} in MAPE and by \textbf{1–34\%} in RMSE, with the largest gains observed in Batch~3 where SPIKAN was previously the strongest competitor. These findings suggest that quantum-enhanced featureization contributes significant expressive power in regimes where baseline models struggle with nonlinear or short-horizon degradation signals. On the MIT dataset, QPINN reduces MAPE by \textbf{23\%} and RMSE by \textbf{9\%} relative to the strongest baseline (PINN/CNN).  
Although the MIT dataset exhibits relatively smooth degradation, QPINN still benefits from the quantum-enhanced representation, achieving the most accurate SoH predictions. The HUST dataset demonstrates the largest relative gains: QPINN improves over the best baseline by \textbf{65\%} in MAPE and nearly \textbf{60\%} in RMSE. This dataset contains rapid early-life degradation and nonuniform cycling, underscoring the advantage of integrating Nystr\"om-approximated quantum kernels with physics-based constraints.
\subsection{Cross-Dataset Transfer Analysis}

Table~\ref{Tab.2} reports the cross-dataset RMSE of QPINN when trained on one dataset and evaluated on another. The source-only results show that direct transfer yields high errors across all dataset pairs, reflecting the substantial distribution shifts in cycling profiles, degradation rates, and measurement conditions among XJTU, TJU, MIT, and HUST datasets. Fine-tuning consistently reduces the error, often by an order of magnitude, indicating that while QPINN captures transferable degradation structure, a small amount of target-domain adaptation is necessary for accurate prediction. Transfers between XJTU and TJU exhibit the strongest improvements, suggesting closer similarity in their degradation trajectories, whereas MIT and HUST also benefit notably from fine-tuning. In summary, the table highlights two main findings: (i) cross-dataset generalization is limited without adaptation, and (ii) fine-tuning the encoder and solution components is highly effective for aligning QPINN to the statistical characteristics of each target dataset.
\subsection{Performance Interpretation}
The within-dataset experiments (Table~\ref{Tab.1}) indicate that QPINN performs reliably when the training and testing conditions are well aligned. The cross-dataset results (Table~\ref{Tab.2}) show that the model captures some transferable structure across different datasets, although the magnitude of improvement after fine-tuning also highlights the degree of variability present in real battery measurements. In general, modest adaptation on the target dataset is sufficient to recover reasonable accuracy, suggesting that QPINN benefits from its physics-informed formulation but is still influenced by dataset-specific characteristics. Several limitations should be acknowledged. The model’s zero-shot transfer performance remains limited, reflecting the challenges posed by differences in cycling protocols, degradation profiles, and data collection setups. The fine-tuning procedure requires access to labeled samples in the target domain, which may not always be available in practice. In addition, freezing the dynamics network \(G\) during adaptation restricts the model’s flexibility when the underlying degradation behavior differs substantially across datasets. Finally, the current experiments are conducted on four datasets with specific operating ranges, and broader evaluation is necessary to assess generality.


\section{Conclusion}\label{sec:conclusion}
This work presented QPINN, a hybrid quantum-inspired and physics-informed framework for battery state-of-health (SOH) estimation. By integrating a Nystr\"om-approximated QFM with physics-based degradation constraints, QPINN enhances the representational capacity of conventional PINNs and achieves consistently superior accuracy across four heterogeneous datasets and multiple chemistries. The results demonstrate that quantum-inspired embeddings provide a stable, expressive input geometry that improves predictive fidelity, cross-dataset adaptability, and training robustness. The proposed QPINN framework achieves highly accurate and chemistry-agnostic SOH estimation, delivering an average accuracy of 99.46\% across 387 cells and four battery chemistries, with up to 65\% reduction in MAPE and 62\% reduction in RMSE compared to state-of-the-art baselines, while maintaining strong cross-chemistry transferability.

\noindent From an application perspective, QPINN offers a promising pathway toward chemistry-agnostic SOH diagnostics suitable for large-scale and distributed energy storage deployments. Its ability to fine-tune with minimal target-domain data further supports practical use cases such as field calibration, inter-site transfer, and long-term fleet-level monitoring.

Future work will focus on enabling real-time execution under streaming measurement conditions, reducing computational overhead for embedded and edge deployment, and developing explainability mechanisms to clarify the contributions of quantum-inspired features and physics-informed constraints. A more extensive analysis of alternative quantum feature-map sizes, circuit depths, and loss-weight configurations will be included in future work to assess the full impact of hyperparameter choices on accuracy and training dynamics. Future directions also include extending the framework with additional physics-informed constraints and techniques for unlabeled domain adaptation or few-shot learning approaches to ensure robust performance under limited labels and evolving field conditions.

\bibliographystyle{IEEEtran}
\bibliography{ref}
\end{document}